\documentclass[journal]{IEEEtran}

\usepackage{graphicx}
\usepackage{amsmath,amssymb}
\usepackage{cite}
\usepackage{url}

\begin{document}

\title{Confidence Geometry Reveals Trace-Level Correctness in Large Language Model Reasoning}

\author{
Shuo Liu\IEEEauthorrefmark{1},
Ding Liu\IEEEauthorrefmark{1},
and Shi-Ju Ran\IEEEauthorrefmark{2}
\thanks{Corresponding authors: Ding Liu (e-mail: liuding@tiangong.edu.cn) and Shi-Ju Ran (e-mail: sjran@cnu.edu.cn).}
\thanks{\IEEEauthorrefmark{1}School of Computer Science and Technology, Tiangong University, Tianjin, China.}
\thanks{\IEEEauthorrefmark{2}Center for Quantum Physics and Intelligent Sciences, Department of Physics, Capital Normal University, Beijing, China.}
}

\maketitle

\begin{abstract}
Large language models (LLMs) generate not only reasoning text, but also token-level confidence trajectories that record how uncertainty evolves during inference. Whether these trajectories are relevant to reasoning correctness remains unclear. Here we show that confidence trajectories encode a content-agnostic confidence geometry associated with trace-level final-answer correctness. Using only token-level confidence values, without access to the input question, reasoning text, hidden states, or external verifiers, we find that low-dimensional representations of confidence trajectories separate correct from incorrect reasoning traces. Across GSM8K, MATH, and MMLU, this geometric separation is quantitatively linked to downstream predictability: stronger clustering of correct and incorrect traces, measured by the Davies--Bouldin index, consistently corresponds to higher correctness-discrimination AUC. We further show that correctness-related information is enriched in the tail of reasoning, suggesting that late-stage confidence dynamics carry key correctness signals. We propose NeuralConf, a lightweight estimator that learns from confidence trajectories for correctness evaluation. Under a fixed trace budget, NeuralConf-derived scores improve confidence-weighted answer aggregation over majority voting, tail confidence, and other static baselines. These results reveal that LLMs expose trace-intrinsic statistical signals of correctness through their own confidence dynamics, offering a route to improve inference using information already present within generation.
Code is available at \url{https://github.com/QML-TGU/NeuralConf}.
\end{abstract}


\section{Introduction}

Large language models (LLMs) have demonstrated impressive performance on tasks requiring complex reasoning, such as mathematics, science, and knowledge-intensive applications \cite{wei2022chain, kojima2022zeroshot, zhou2023least}. Despite their success, the reliability of reasoning in LLMs remains a major challenge. A model can generate a seemingly plausible chain of thought and still produce incorrect answers, either due to early local errors that propagate through the reasoning process or because the reasoning trajectory, while coherent, leads to the wrong conclusion. To address this, several approaches have proposed aggregating multiple reasoning traces to improve answer robustness \cite{wang2023self, yao2023tree, snell2025scaling, brown2024large}. However, these methods do not provide insight into why some reasoning traces succeed while others fail, leaving a critical open question on what distinguishes successful reasoning from unsuccessful reasoning in LLMs.

Existing approaches generally focus on analyzing the content of the reasoning trace or applying additional supervisory signals. External verifiers, process-supervised models, and self-evaluation methods assess the quality of a reasoning trace based on its linguistic content, intermediate steps, or through an additional checking procedure \cite{cobbe2021training, lightman2024let, weng2023selfverification, xie2023selfeval, jacovi-etal-2024-chain}. Another parallel line of work has explored how model confidence and uncertainty can serve as useful signals for predicting answer correctness and improving aggregation \cite{kadavath2022language, kuhn2023semantic, xiong2024confidence, lin2022teaching}. However, existing methods do not address whether the internal statistical information during reasoning reflects the structure that determines eventual correctness.

Here we study whether token-level confidence trajectories contain trace-level correctness information that can be learned without access to reasoning content. We introduce NeuralConf, a lightweight estimator that learns from confidence trajectories alone and produces a trace-level correctness score. We restrict the input of NeuralConf to the one-dimensional confidence sequence observed during decoding, and do not use the input question, reasoning text, hidden states \cite{azaria2023internal, chen2024inside}, external verifiers, or auxiliary knowledge sources. Under this constraint, any recoverable signal must originate from the confidence trajectory itself rather than from semantic content or post hoc verification.

We find that confidence trajectories encode a measurable content-agnostic confidence geometry of reasoning correctness. Across GSM8K, MATH, and MMLU, low-dimensional representations derived from confidence trajectories separate correct from incorrect traces, revealing a clustering structure associated with final-answer correctness. Clustering quality, measured by the Davies--Bouldin index, is consistently linked to correctness-discrimination AUC: stronger separation between correct and incorrect traces corresponds to higher AUC. We further find that this structure is unevenly distributed along the trajectory and becomes more recoverable when longer tail-aligned segments are observed, indicating that late-stage confidence dynamics carry especially informative signals about whether a trace will end correctly.

This view differs from existing confidence-based approaches that collapse token-level confidence into fixed summaries such as tail confidence or grouped confidence~\cite{fu2026deep}. Such summaries assume in advance where and how correctness-related information should appear. NeuralConf instead learns the relevant structure from the temporal trajectory itself. As a functional test, NeuralConf-derived scores improve confidence-weighted answer aggregation over majority voting, tail confidence, and other static confidence baselines under a fixed trace budget on multiple benchmarks. Our work shows that LLM reasoning traces contain trace-intrinsic, content-agnostic statistical structure associated with correctness, and that this structure can be used to improve inference using information already present within the generation process.

\begin{figure*}[t]
    \centering
    \includegraphics[width=\textwidth]{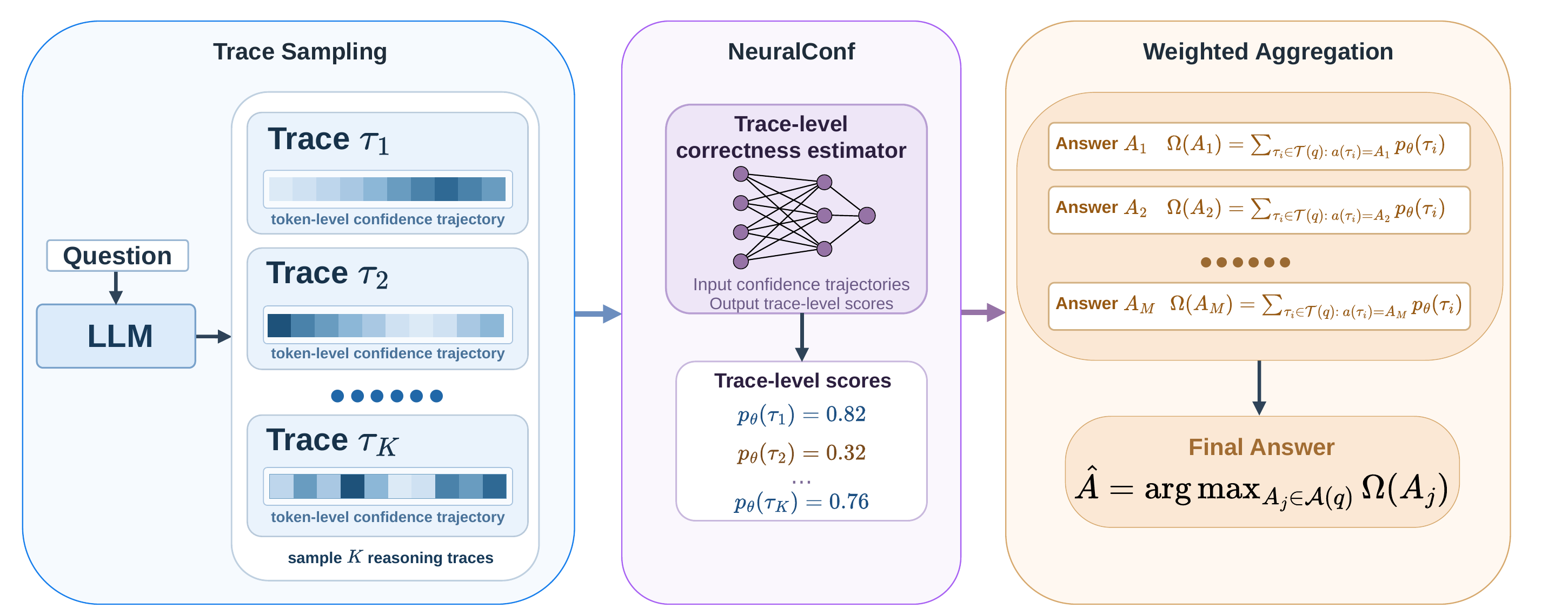}
    \caption{
    Overview of the confidence-only readout protocol. A frozen backbone LLM samples multiple reasoning traces for each question. Each trace produces a final answer and a token-level confidence trajectory. NeuralConf observes only the confidence trajectory and maps it to a trace-level score. The learned representation is used to analyze confidence-trajectory geometry, while the scalar score is used for trace-level discrimination and answer aggregation under a fixed trace budget.
    }
    \label{fig:overview}
\end{figure*}

\section{Method}
\label{sec:method}

We study whether trace-level final-answer correctness is recoverable from the
confidence dynamics generated during reasoning. The central design principle is
isolation: the readout observes only a one-dimensional token-level confidence
trajectory of a sampled reasoning trace. It does not observe the input question,
the generated reasoning text, model hidden states, verifier outputs, or the
final-answer string during scoring. Under this restriction, any recoverable signal
must originate from the confidence dynamics themselves rather than from semantic
content or post hoc verification.

For each question \(q\), a frozen backbone LLM samples a set of reasoning traces
\[
\mathcal{T}(q)=\{\tau_1,\ldots,\tau_K\}.
\]
Each trace \(\tau\) terminates in a final answer \(a(\tau)\) and has length
\(L_\tau\).
\begin{figure*}[t]
    \centering
\includegraphics[width=0.97\textwidth]{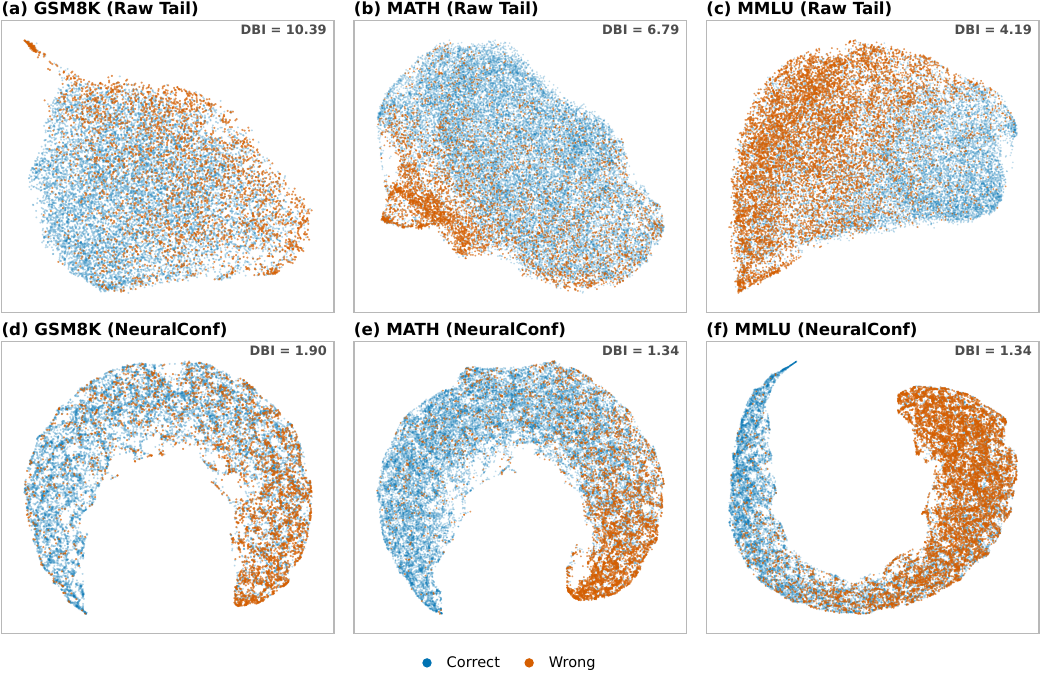}
\caption{
Representation-level evidence that confidence trajectories contain correctness-related structure. UMAP~\cite{mcinnes2018umap} visualizations of raw confidence trajectories and NeuralConf embeddings, derived from the same final 2048-token segment and the same trace subset, are shown for GSM8K, MATH and MMLU. In each case, the learned representation separates correct and incorrect traces more clearly than the raw confidence sequence. Colours indicate trace-level final-answer correctness, and DBI~\cite{davies1979cluster} is computed in the original feature space before projection.
}
\label{fig:representation}
\end{figure*}
\subsection{Confidence trajectory construction}
\label{sec:confidence_construction}

At each decoding position \(i\), we compute a token confidence score following
DeepConf. Let \(P_i(j)\) denote the probability of the \(j\)-th highest-probability
token among the top-\(k\) candidates at position
\(i\). The token confidence is defined as
\[
c_i = -\frac{1}{k}\sum_{j=1}^{k}\log P_i(j),
\]
where \(k=20\) in all experiments.

This score is a top-\(k\) distributional peakedness measure rather than the
negative log-likelihood of the sampled token. When the output distribution is
more concentrated, lower-ranked top probabilities \(k\) tend to become smaller,
which increases their negative log-probabilities. Following DeepConf, larger
\(c_i\) is therefore interpreted as higher token confidence, while smaller
\(c_i\) indicates greater uncertainty in local token prediction.

The generated trace \(\tau\) is then represented by its token-level confidence
trajectory
\[
\mathbf{c}(\tau)=(c_1,c_2,\ldots,c_{L_\tau}).
\]

A binary correctness label \(y(\tau)\in\{0,1\}\) is assigned by comparing the final
answer \(a(\tau)\) with the ground truth. This label is used only for training and
evaluation, not as an input to the readout.

The same token-level confidence definition is used for NeuralConf and for all
DeepConf-style hand-crafted baselines, including TailConf and Bottom-10Conf.
Therefore, our comparisons isolate how the resulting confidence trajectory is
modeled and aggregated, rather than how the underlying token-level confidence
signal is constructed.
\subsection{Tail-aligned confidence trajectories}

Reasoning traces have variable length, so we convert each confidence trajectory into a fixed-size tail-aligned segment of length \(L_{\max}\). If a trajectory is longer than \(L_{\max}\), only its final \(L_{\max}\) confidence values are retained. If it is shorter, it is left-padded and paired with a binary mask so that padded positions do not contribute to pooled representations. We denote the resulting fixed-size representation by
\[
\tilde{\mathbf{c}}(\tau)\in\mathbb{R}^{L_{\max}}.
\]

Tail alignment has two roles. First, it provides a uniform representation for variable-length traces. Second, it focuses the readout on the late-stage confidence dynamics closest to answer commitment, where correctness-related information may be most concentrated. By varying \(L_{\max}\), we test whether the recoverable signal is exhausted by short summaries or becomes more resolvable as longer portions of the trajectory are observed. In addition, fixed-size sliding windows are used in later analyses to separate the effect of trajectory position from the effect of observed length.

\subsection{NeuralConf as a confidence-trajectory readout}

NeuralConf is a lightweight sequence readout from confidence trajectories to trace-level correctness scores:
\[
p_\theta(\tau)=f_\theta\!\left(\tilde{\mathbf{c}}(\tau)\right)\in(0,1).
\]
The readout is intentionally restricted to the one-dimensional confidence sequence. It is not given the question, the reasoning text, hidden states, external verifier scores, or the final-answer string. Thus, NeuralConf should be understood not as a semantic verifier, but as an instrument for testing whether confidence trajectories contain learnable correctness-related structure.

In our implementation, \(f_\theta\) consists of a compact one-dimensional convolutional encoder with residual blocks~\cite{he2016deep}, followed by a small prediction head. The encoder produces a trace representation \(\mathbf{h}(\tau)\), and the prediction head maps this representation to the scalar score \(p_\theta(\tau)\). We use \(\mathbf{h}(\tau)\) for representation-level analysis and \(p_\theta(\tau)\) for trace-level discrimination and answer aggregation. 

NeuralConf is trained using only binary final-answer correctness labels. In practice, we optimize a class-weighted binary cross-entropy objective, with class weights used to compensate for imbalance between correct and incorrect traces~\cite{he2009imbalanced}. All train, validation, and test splits are performed at the question level rather than the trace level, so that traces sampled from the same question never appear in more than one split. This prevents leakage through question-specific confidence patterns and ensures that evaluation measures generalization to unseen questions.

\subsection{Geometry and trace-level discrimination}

The learned representation is used to examine whether confidence trajectories induce a geometry associated with correctness. For each held-out trace, NeuralConf maps the tail-aligned confidence trajectory to an embedding \(\mathbf{h}(\tau)\). We then compare the organization of correct and incorrect traces in the learned representation space with their organization in the raw confidence-sequence space.

This analysis is deliberately representation-level rather than answer-level. The goal is to test whether correctness-related structure is present in confidence trajectories before using final-answer strings for aggregation. We quantify this structure using the Davies--Bouldin index (DBI), where lower values indicate stronger separation between correct and incorrect traces. We also evaluate the scalar score \(p_\theta(\tau)\) as a trace-level discriminator using AUC. Together, DBI and AUC test whether the same confidence-only signal is geometrically organized and predictively useful.

\subsection{Confidence-weighted answer aggregation}

Finally, we test whether the confidence-trajectory signal is functionally relevant for multi-trace inference. For a question \(q\), the backbone model samples \(K\) traces. NeuralConf first scores each trace independently using only its confidence trajectory. The final-answer string is used only after scoring, when traces are grouped by their predicted answers.

Let \(\mathcal{A}(q)=\{A_1,\ldots,A_M\}\) denote the set of distinct candidate answers produced by the sampled traces. Each candidate answer receives the summed NeuralConf score of all traces that terminate in that answer:
\[
\Omega(A_j)
=
\sum_{\tau_i\in\mathcal{T}(q):\,a(\tau_i)=A_j}
p_\theta(\tau_i).
\]
The final prediction is
\[
\hat{A}
=
\arg\max_{A_j\in\mathcal{A}(q)}
\Omega(A_j).
\]

When all trace scores are equal, this rule reduces to majority voting. The aggregation experiment therefore serves as a downstream functional test: if confidence trajectories contain correctness-related information beyond static confidence summaries, then scores derived from those trajectories should improve answer selection under the same fixed trace budget.

\section{Experiments}

We evaluate NeuralConf in two roles: as a readout of correctness-related structure in confidence trajectories, and as a trace-scoring signal for answer aggregation. The experiments were organized around five questions. Do confidence trajectories contain correctness-related structure that a learned estimator can recover more clearly than is apparent in the raw sequences themselves? Does this structure become more resolvable as longer trajectory segments are observed? Is the informative signal uniformly distributed along the trajectory, or preferentially concentrated near particular stages of reasoning? If such structure is real, does it have downstream relevance for answer aggregation under a fixed trace budget? And finally, how does the strength of this signal change in a harder reasoning regime and at larger model scale?

\subsection{Experimental design}

We evaluated the method on three main reasoning benchmarks: GSM8K \cite{cobbe2021training}, MATH \cite{hendrycks2021math} and MMLU \cite{hendrycks2021mmlu}. For each benchmark, we constructed a 1,200-question subset and sampled $K=128$ chain-of-thought reasoning traces per question from a frozen backbone model. Questions, rather than traces, were split into training, validation and test partitions of 600, 300 and 300, respectively, and all principal representation and aggregation results were measured on the held-out test split. For MMLU, we used a subset of college- and high-school-level STEM subjects; the full subject list is provided in Appendix E. In a final boundary-case analysis, we additionally examine ReClor \cite{yu2020reclor}, a harder single-answer multiple-choice benchmark with a small answer space and strong distractors.

We used DEEPSEEK-R1-DISTILL-QWEN-1.5B as the backbone model for the main three-benchmark study \cite{deepseek2025r1,yang2024qwen25,yang2024qwen25math}. For the ReClor boundary-case analysis, we compared 1.5B and 7B backbones under the same trajectory construction, input-length sweep and NeuralConf estimator pipeline. Along each sampled trace, we computed a token-level confidence score at every decoding step from the model's top-$k$ output probabilities, thereby obtaining the confidence trajectory used throughout this study. NeuralConf received only this one-dimensional trajectory as input. Unless otherwise stated, reported results are based on tail-aligned inputs. For analyses of input length, we varied the maximum trajectory length over
\[
L_{\max}\in\{4,8,16,32,64,128,256,512,1024,2048\},
\]
whereas for the main aggregation experiments we used $L_{\max}=2048$. If a trace exceeded $L_{\max}$, only its final $L_{\max}$ confidence values were retained; otherwise, the sequence was left-padded with zeros and paired with a binary mask so that padded positions did not contribute to the pooled representation.

Our analysis uses two complementary classes of metrics. To assess the geometry of the learned representation, we use the Davies--Bouldin index (DBI) \cite{davies1979cluster}, with lower values indicating better class separation. To assess trace-level discrimination, we use the area under the receiver operating characteristic curve (AUC) \cite{fawcett2006roc}. These representation-level measures are complemented by answer-level aggregation accuracy, which serves as a functional test of whether the signal extracted from confidence trajectories is sufficiently informative to alter final answer selection.
\begin{figure*}[!t]
    \centering
    \includegraphics[width=0.97\textwidth]{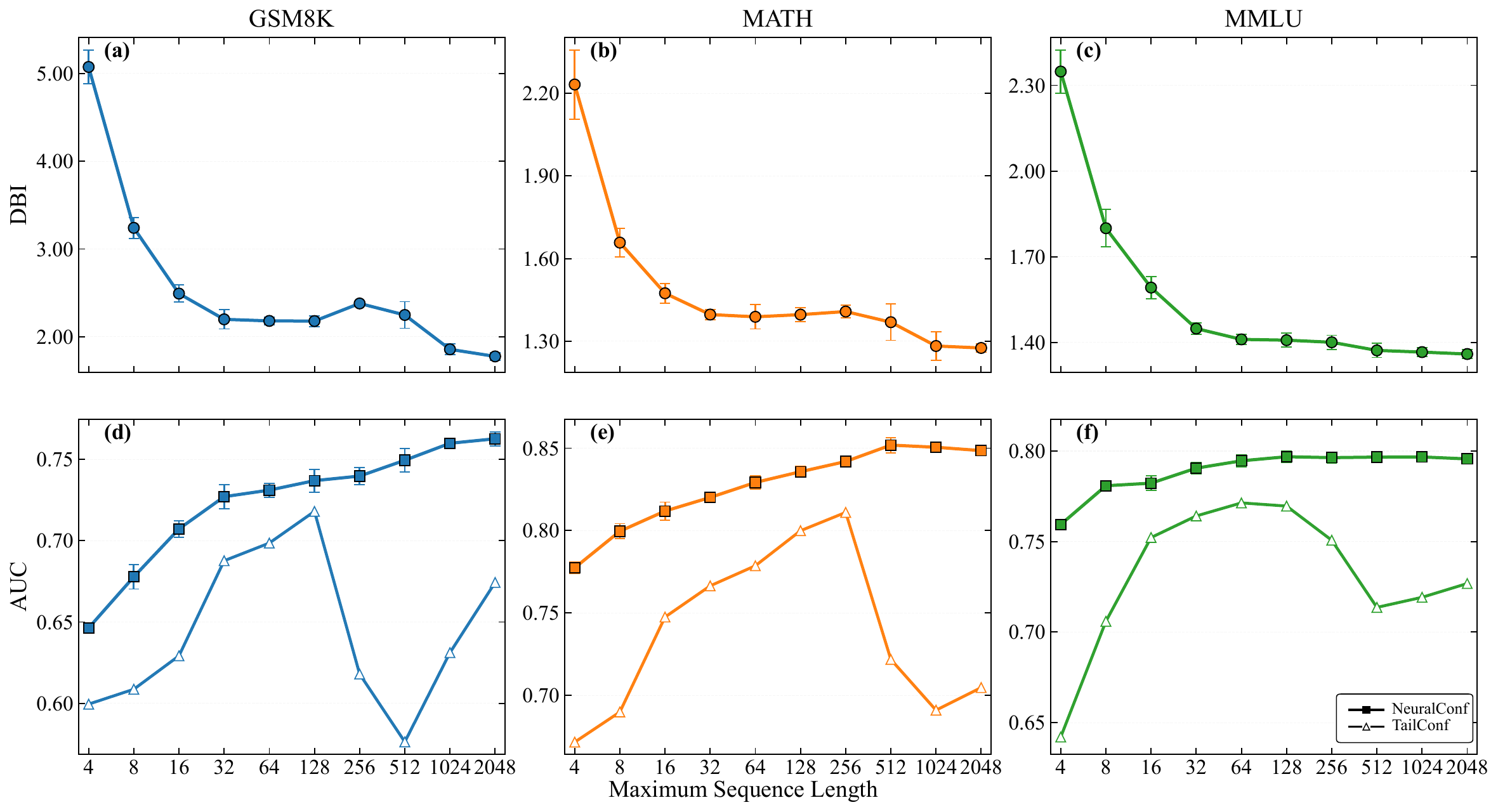}
   \caption{
Dependence of representation quality and trace-level discrimination on the maximum input length. Top, DBI of NeuralConf embeddings across input lengths on GSM8K, MATH and MMLU, with lower values indicating better separation between correct and incorrect traces. Bottom, trace-level AUC for NeuralConf and TailConf across the same length settings. Error bars are shown for NeuralConf only; TailConf is deterministic for a fixed trace set and therefore has none. For NeuralConf, error bars denote mean $\pm$ s.d. over five random seeds.
}
    \label{fig:length}
\end{figure*}
\subsection{Confidence trajectories contain learnable correctness-related structure}

Confidence trajectories contain correctness-related structure, and NeuralConf exposes it more clearly than the raw sequences do. Across GSM8K, MATH and MMLU, representations learned by NeuralConf separated traces with correct and incorrect final answers more clearly than the corresponding raw confidence trajectories (Fig.~\ref{fig:representation}). For a controlled comparison, we restrict this analysis to traces with confidence length at least 2048 and use the same trace subset for both the raw and learned representations. The raw representation is given by the final 2048 confidence values of each trace, whereas the learned representation is the NeuralConf encoder embedding derived from that same tail-2048 segment. The key point is not that supervision improves separation, but that this separation is recovered from confidence trajectories alone. The input to NeuralConf is a one-dimensional confidence trajectory, with no access to reasoning text or hidden states. The learned embedding thus provides evidence that token-level confidence trajectories carry structured information associated with eventual correctness, rather than merely constituting incidental fluctuations around a final answer.

\subsection{Longer tail segments reveal progressively more structure}

The signal becomes easier to recover as longer tail segments are observed. As increasingly long tail-aligned segments were provided, the learned representation generally became more separable across all three benchmarks, as reflected by a progressive decline in DBI and a corresponding increase in trace-level AUC (Fig.~\ref{fig:length}). This pattern indicates that the informative structure is not exhausted by a short local summary, but becomes progressively more resolvable as additional portions of the trajectory are made visible.

This analysis is particularly informative when compared with TailConf, which is shown in the main text as the most direct tail-based fixed-summary baseline. If the correctness-related information contained in confidence trajectories were already well approximated by such a summary, then a learned sequence model would be expected to offer only limited and unstable gains once that summary had access to a sufficiently large tail segment. Instead, NeuralConf maintains a consistent AUC advantage over TailConf across the full range of tested input lengths. The implication is not simply that learned modelling performs better, but that the relevant signal is not fully captured by a fixed summary. In other words, useful information about correctness appears to be distributed across the trajectory in a form that is only partially recoverable by static summarization.

A complementary grouping-length analysis of Bottom-10Conf is provided in Appendix A. Notably, even when Bottom-10Conf is computed over the full available confidence trajectory and tuned over grouping length, its strongest trace-level AUC remains below that achieved by NeuralConf at the longest tail-aligned setting used in the main text. This strengthens the interpretation that the advantage of the learned estimator is not tied to a single hand-crafted comparator, but reflects a broader limit of fixed confidence compression.

A further point emerges from the different shapes of the length curves. TailConf often improves only up to moderate input lengths and can flatten or deteriorate thereafter, whereas NeuralConf benefits more steadily from longer tail-aligned inputs before gradually saturating. Taken together, these results support the view that correctness-related information in token-level confidence trajectories is cumulative and sequence-dependent, rather than reducible to a small set of fixed summary statistics near the end of the trace.

\begin{figure*}[!t]
    \centering
    \includegraphics[width=0.97\textwidth]{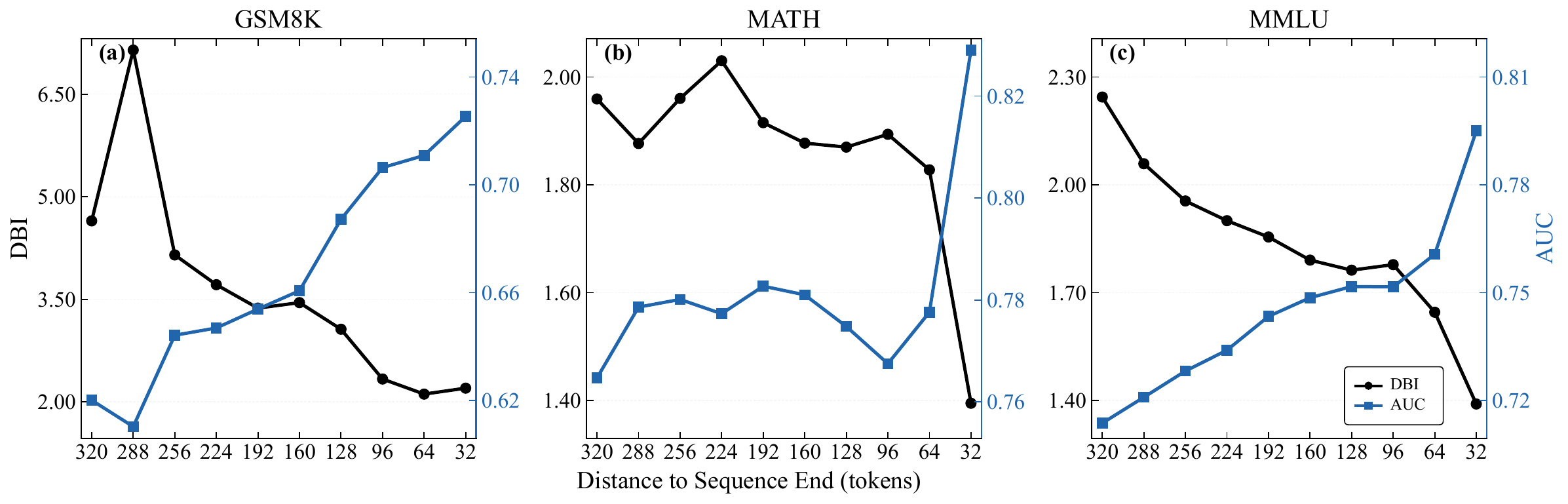}
    \caption{
    Positional organization of correctness-related structure along the confidence trajectory. DBI and trace-level AUC are shown as functions of window position using a fixed sliding window of size 64 and stride 32. The horizontal axis denotes the distance between the window centre and the end of the sequence. Across all three benchmarks, windows nearer the end of the trajectory generally provide stronger discrimination.
    }
    \label{fig:position}
\end{figure*}
\subsection{Correctness-related structure is concentrated near the end of reasoning}

To disentangle position from length, we next use a fixed-size sliding window. The analysis shows that the signal is not uniformly distributed along the trajectory. Across GSM8K, MATH and MMLU, windows located closer to the end of the sequence generally yielded lower DBI and higher trace-level AUC than windows located further away from the end (Fig.~\ref{fig:position}). The simplest interpretation is that correctness-related confidence structure is not evenly distributed across reasoning, but becomes more concentrated near the final stages of the trace. Late-stage portions of the confidence trajectory may therefore more directly reflect whether a reasoning path is approaching a correct or incorrect final answer.

The sliding-window curves also show local irregularities, which likely reflect the fact that short cropped windows preserve only a limited segment of the trajectory and may cut across transitions between different phases of reasoning. Nevertheless, because the window size is held fixed throughout, the global trend is unlikely to be explained merely by access to more tokens. Rather, it indicates that the informative content of confidence trajectories is position-dependent and preferentially enriched near the tail of reasoning.

\subsection{Functional relevance for answer aggregation}

NeuralConf yields the highest aggregation accuracy on all three benchmarks under a fixed trace budget (Table~\ref{tab:aggregation}). We therefore asked whether the trace-level signal recovered from confidence trajectories also matters downstream at the level of answer aggregation. To test this, we compared NeuralConf with majority voting and with DeepConf-style hand-crafted baselines using the same fixed budget of 128 traces per question. For the hand-crafted baselines, we followed the configurations recommended by DeepConf\cite{fu2026deep}: TailConf was computed from the final 2048 token-level confidence values of each trace, whereas Bottom-10Conf used grouping length 1024.

The gains are modest but consistent. Notably, limited answer-level movement was not unique to NeuralConf: under a fixed budget of 128 traces per question, the DeepConf-style hand-crafted baselines also changed majority voting only marginally, except when aggressive 10\% filtering substantially reduced the available voting set. This suggests that answer-level aggregation is a comparatively coarse endpoint of trace quality in the present setting. The same internal signal that yields clearer representation-level separation and higher trace-level AUC also improves downstream answer selection, but aggregation should therefore be interpreted as a functional assay of relevance rather than as the most sensitive endpoint of the phenomenon. Within this regime, NeuralConf remains the strongest method among those tested.

This interpretation is important for comparison with DeepConf. The original DeepConf offline voting results were reported at a substantially larger trace budget (\(K=512\)), with 10\% filtering retaining roughly 51 traces for final voting, whereas the present study uses \(K=128\). We therefore interpret the weaker behaviour of aggressive filtering here as regime-dependent rather than as evidence against the underlying idea. Under the reduced budget considered here, continuous weighting appears more robust because it preserves a larger evidential base for answer-level aggregation.

Taken together, these aggregation results show that the information carried by confidence trajectories is not only decodable at the level of individual traces, but also consequential for final answer selection. Although answer-level aggregation is a coarser endpoint than representation-level separation or trace-level AUC, its improvement indicates that the internal organization captured by NeuralConf is functionally relevant to multi-trace reasoning under a fixed generation budget.

\begin{table*}[!t]
    \centering
\caption{
Answer-aggregation accuracy under a fixed budget of 128 sampled reasoning traces per question. NeuralConf results are reported as mean $\pm$ s.d. over five random seeds; all other methods are deterministic for the fixed trace set.
}
    \label{tab:aggregation}
    \begin{tabular}{lccc}
        \hline
        Method & GSM8K & MATH & MMLU \\
        \hline
        Majority Voting & 0.9333 & 0.9467 & 0.8167 \\
        Bottom-10Conf (10\%) & 0.9200 & 0.9267 & 0.7333 \\
        Bottom-10Conf (90\%) & 0.9367 & 0.9467 & 0.8133 \\
        TailConf (10\%) & 0.9233 & 0.9367 & 0.7467 \\
        TailConf (90\%) & 0.9367 & 0.9467 & 0.8200 \\
        \textbf{NeuralConf} & $\mathbf{0.9407 \pm 0.0015}$ & $\mathbf{0.9493 \pm 0.0015}$ & $\mathbf{0.8240 \pm 0.0028}$ \\
                \hline
    \end{tabular}
\end{table*}

\subsection{ReClor reveals a scale-dependent boundary case}

ReClor shows a different recovery profile from GSM8K, MATH and MMLU. 
Whereas those benchmarks exhibit a largely monotonic benefit from observing longer tail segments, ReClor shows only limited gains from additional tail length and a much stronger dependence on model scale. 
This benchmark is particularly stringent because it is single-answer multiple-choice, with a small answer space and strong distractors, so an internally coherent trace can still terminate in the wrong option.

\begin{figure}[t]
    \centering
    \includegraphics[width=\linewidth]{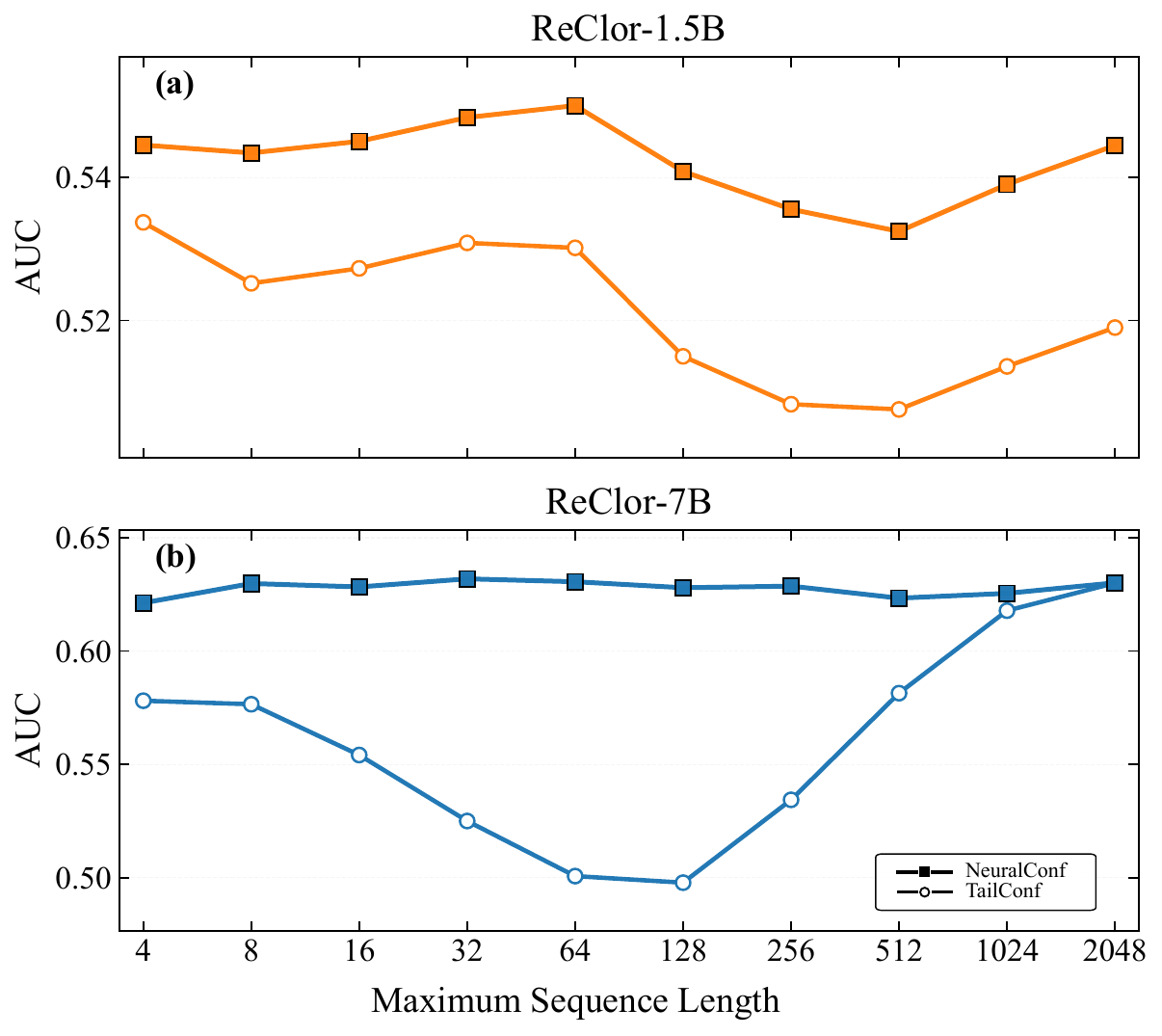}
   \caption{
    Scale-dependent recoverability of correctness-related signals on ReClor. 
    Trace-level AUC is shown as a function of maximum sequence length for 1.5B (a) and 7B (b) backbones, comparing NeuralConf and TailConf. 
    At 1.5B, both methods remain only modestly above chance across sequence lengths. 
    At 7B, NeuralConf maintains consistently higher AUC, whereas TailConf shows substantially greater sensitivity to input length and approaches NeuralConf only at the longest lengths.
    }
    \label{fig:reclor_scale}
\end{figure}

At 1.5B, both methods remain only modestly above chance across the full length sweep, indicating weak coupling between confidence trajectories and final-answer correctness in this regime. 
NeuralConf nevertheless stays consistently above TailConf. 
At 7B, recoverability improves substantially: NeuralConf remains high and comparatively stable across almost the entire range, whereas TailConf is strongly length-sensitive, dropping at intermediate lengths and approaching NeuralConf only when the longest tail segments are retained (Fig.~\ref{fig:reclor_scale}).

These results refine the scope of the main claim. 
Confidence trajectories continue to encode information about trace correctness on ReClor, but that information is less robust than on the other benchmarks and is not recovered simply by exposing longer tail segments. 
Instead, successful recovery depends more strongly on model scale and on preserving trajectory structure through sequence-level modelling.

\section{Conclusion}

We set out to test a basic but unresolved question: whether token-level confidence trajectories generated during LLM reasoning contain learnable structure associated with trace-level final-answer correctness, beyond what can be captured by fixed confidence summaries. Our results show that they do. Across GSM8K, MATH and MMLU, confidence trajectories alone support a learned representation in which correct and incorrect traces become more clearly separable than in the raw sequence space. More importantly, this signal is not merely present but organized: correctness-related information is distributed along the trajectory, becomes increasingly recoverable as longer tail-aligned segments are observed, and is preferentially concentrated near the end of reasoning. Scores derived from these trajectories also improve downstream answer aggregation under a fixed trace budget, showing that the internal structure captured by NeuralConf is not only measurable at the level of individual traces, but also useful for final answer selection. Taken together, these results support a broader view of confidence in LLM reasoning: it is more informatively understood as a sequence-resolved internal trajectory than as a static scalar attached to an answer.

ReClor adds an important second layer to this claim. In this harder reasoning regime, where a single correct option must be selected from a small set of strong alternatives, correctness-related structure remains present in confidence trajectories, but its recoverability depends more strongly on model scale. At 1.5B, the signal is only weakly recoverable; at 7B, it becomes substantially more recoverable. Across both scales, NeuralConf remains more robust than TailConf across input lengths, indicating that learned sequence-level modelling is better able to preserve and capture this structure when the task becomes more demanding. ReClor thus shows that the recoverability of correctness-related information is not governed by tail length alone, but varies across reasoning regimes and models.

A central task for future work is therefore not simply to improve trace scoring further, but to characterize the scope and form of this phenomenon itself: when confidence trajectories become informative, where along the reasoning process that information is concentrated, and why its recoverability varies across tasks, models and answer spaces. By identifying learnable organization within internal confidence trajectories, this work highlights an underused level of structure in LLM reasoning and suggests a broader framework for studying reasoning reliability beyond the final text alone.
\section*{Acknowledgment}

A portion of the numerical simulations was carried out using the robotic AI-Scientist platform of the Chinese Academy of Sciences. This work was supported in part by the Tianjin Natural Science Foundation of China under Grant No. 20JCYBJC00500 and by the Science and Technology Development Fund for Higher Education of Tianjin Education Commission under Grant No. 2018KJ217.

\appendices
\section{Grouping-Length Dependence of Bottom-10Conf}
\label{app:bottom10_group}

Unlike TailConf, which is naturally aligned with the tail-length sweep used in the main text, Bottom-10Conf is defined through grouped summaries of the confidence trajectory and is therefore governed by grouping granularity rather than by tail length alone. We therefore analyse Bottom-10Conf separately as a function of grouping length, rather than forcing it into the matched tail-length comparison used for NeuralConf and TailConf.

\begin{figure*}[!t]
    \centering
    \includegraphics[width=0.97\textwidth]{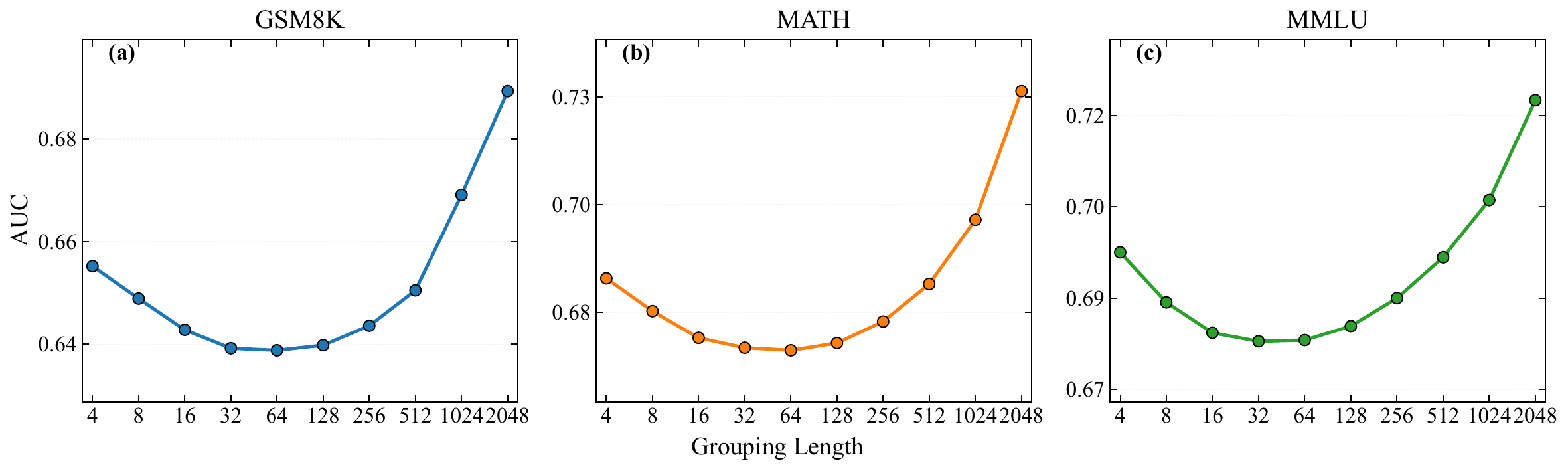}
    \caption{
    Trace-level AUC of Bottom-10Conf as a function of grouping length on GSM8K, MATH and MMLU. Bottom-10Conf is computed from the full available confidence trajectory using the same grouped-summary formulation as in the aggregation experiments, while varying the grouping length from 4 to 2048. 
    }
    \label{fig:bottom10_grouping}
\end{figure*}

For each trace, we compute Bottom-10Conf from the full available confidence trajectory using the same grouped-summary formulation as in the aggregation experiments, while varying the grouping length over
\[
\{4,8,16,32,64,128,256,512,1024,2048\}.
\]
Across GSM8K, MATH and MMLU, Bottom-10Conf exhibits a clear dependence on grouping length, with trace-level discrimination initially decreasing at short-to-moderate grouping lengths and then increasing again as the grouping length becomes larger (Fig.~\ref{fig:bottom10_grouping}).

However, the more important observation is comparative rather than configurational. Even at its strongest grouping-length setting, Bottom-10Conf remains below the trace-level AUC achieved by NeuralConf at the longest tail-aligned setting used in the main text. This indicates that the advantage of the learned estimator cannot be reduced to a favourable choice of hand-crafted hyperparameters. Rather, it supports the broader interpretation advanced throughout the paper: although grouped confidence summaries recover some correctness-related information from the trajectory, they do so only partially, and their performance remains more sensitive to design choice than that of the learned sequence-level estimator.

The non-monotonic shape of the Bottom-10Conf curves is itself informative. Unlike NeuralConf, whose discrimination improves more steadily as increasingly long tail segments are made available, Bottom-10Conf depends strongly on how the trajectory is partitioned. This sensitivity is consistent with the view that correctness-related structure is distributed across the confidence sequence in a way that is not fully captured by a fixed grouped summary.

\section{Distribution-Level Evidence for Stronger Class Separation}
\label{app:score_dist}

The main text establishes that NeuralConf yields lower DBI and higher AUC than hand-crafted confidence summaries. To complement these representation- and ranking-based results, we further examine the full score distributions assigned to correct and incorrect traces on the held-out test set.

\begin{figure*}[!t]
    \centering
    \includegraphics[width=0.97\textwidth]{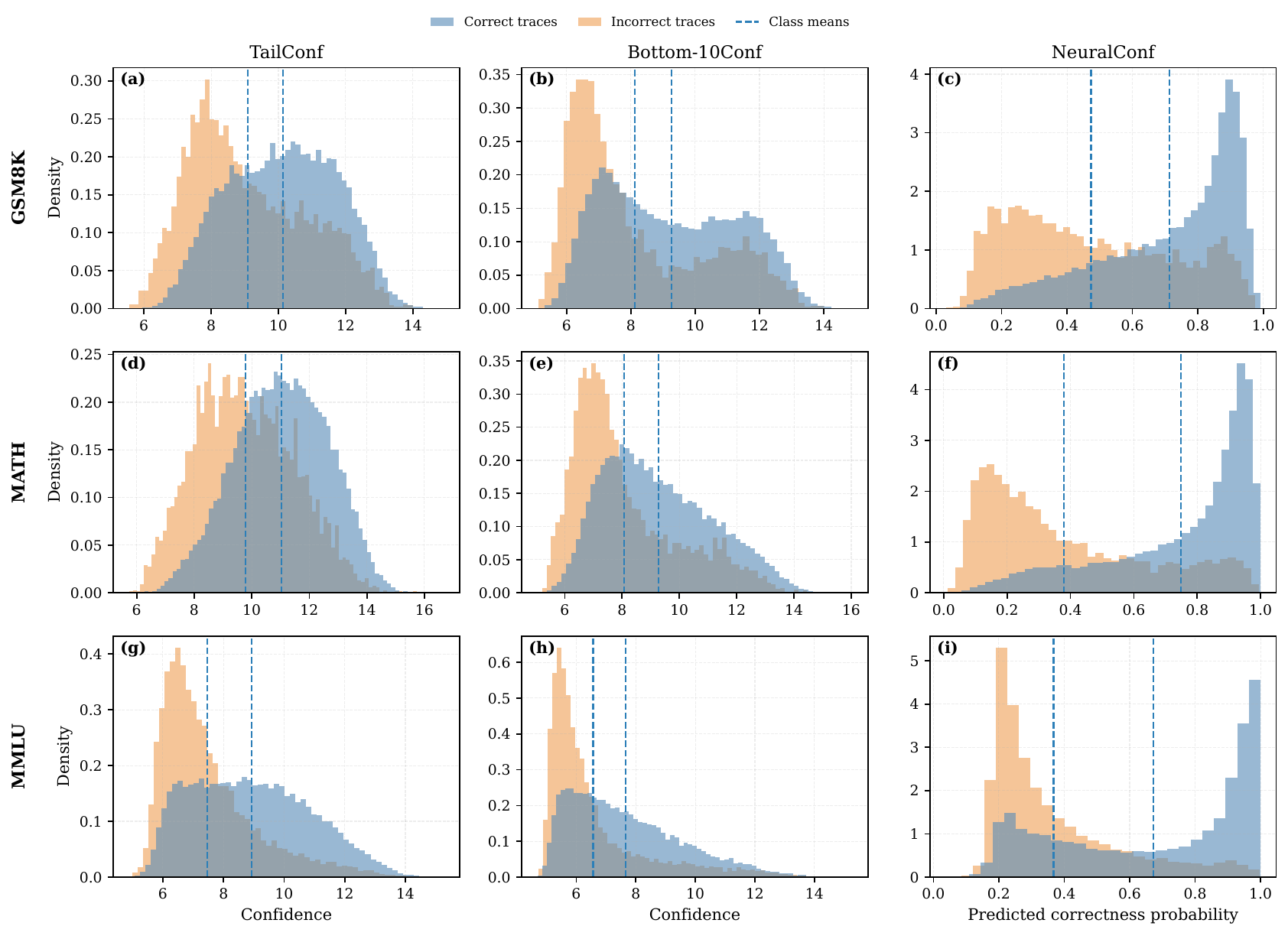}
    \caption{
    Distributional comparison of trace-level scores for incorrect and correct traces on GSM8K, MATH and MMLU. Rows correspond to datasets and columns correspond to TailConf, Bottom-10Conf and NeuralConf. Histograms are computed on held-out test traces, and dashed vertical lines denote class means. Across all three datasets, NeuralConf produces a more distinct distributional gap between correct and incorrect traces than the hand-crafted baselines.
    }
    \label{fig:score_distribution}
\end{figure*}

Across GSM8K, MATH and MMLU, NeuralConf produces a clearer distributional gap between correct and incorrect traces than either TailConf or Bottom-10Conf (Fig.~\ref{fig:score_distribution}). In particular, the NeuralConf readout shifts a larger fraction of correct traces toward high-score regions while simultaneously pushing more incorrect traces toward lower-score regions. This pattern is consistent with the central interpretation of the paper: the learned readout does not merely alter average scores, but recovers a more structured decision-relevant organization from the confidence trajectory itself.

By contrast, the hand-crafted baselines show substantially greater overlap between the two classes. This observation is important because it reinforces the claim that the advantage of NeuralConf is not reducible to a minor calibration effect. Rather, the learned model appears to recover information about correctness that is distributed across the trajectory and only partially accessible to fixed summaries.

\section{Additional Threshold-Dependent Accuracy Analysis}
\label{app:trace_acc}

The main text focuses on DBI and AUC because they directly characterize the geometry of the learned representation and the ranking quality of the trace-level readout. As a complementary analysis, we also report trace-level classification accuracy under a fixed decision threshold.

\begin{figure*}[!t]
    \centering
    \includegraphics[width=0.97\textwidth]{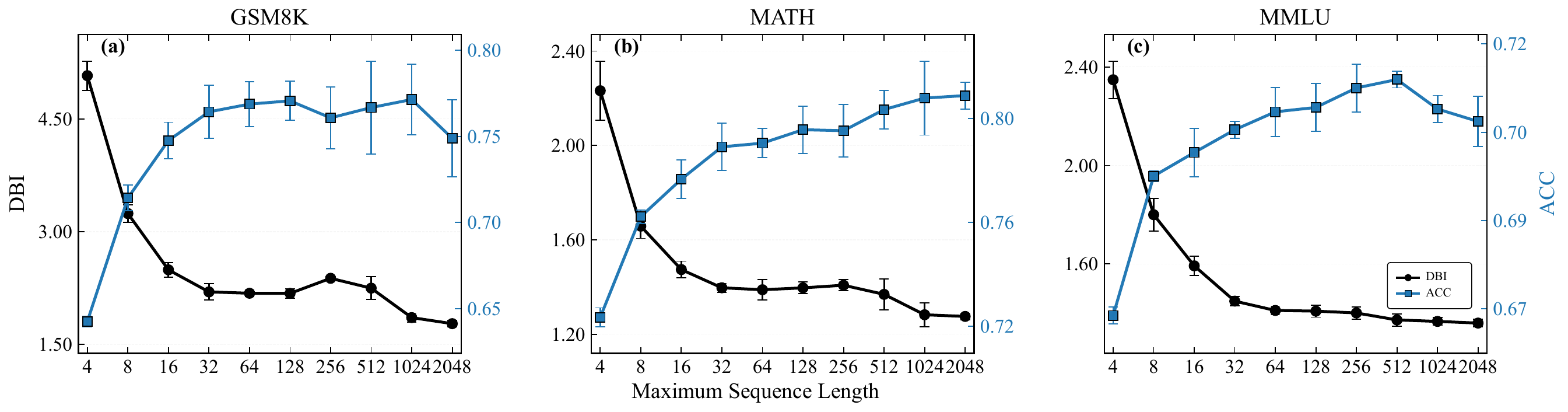}
    \caption{
    Trace-level accuracy as a function of the maximum input length on GSM8K, MATH and MMLU, shown together with DBI for reference. Results are averaged over five random seeds. Although threshold-dependent, the accuracy trends are broadly consistent with the AUC analyses reported in the main text.
    }
    \label{fig:trace_accuracy}
\end{figure*}

Across GSM8K, MATH and MMLU, trace-level accuracy generally increases as longer tail-aligned trajectory segments are made available (Fig.~\ref{fig:trace_accuracy}). Although threshold-dependent accuracy is less informative than AUC for imbalanced settings, it provides an additional view of how the learned signal strengthens with increasing input length. The same broad pattern observed in the main text reappears here: longer tail-aligned inputs support more reliable discrimination between traces that terminate correctly and those that do not.

This analysis should be interpreted as supplementary rather than primary. Because accuracy depends on a particular threshold, it is not the most sensitive measure of class separation. Nonetheless, the consistency between the accuracy trends here and the AUC trends in the main text supports the robustness of the conclusion that correctness-related information becomes progressively more readable as additional portions of the trajectory are observed.

\begin{figure*}[!t]
    \centering
    \includegraphics[width=0.97\textwidth]{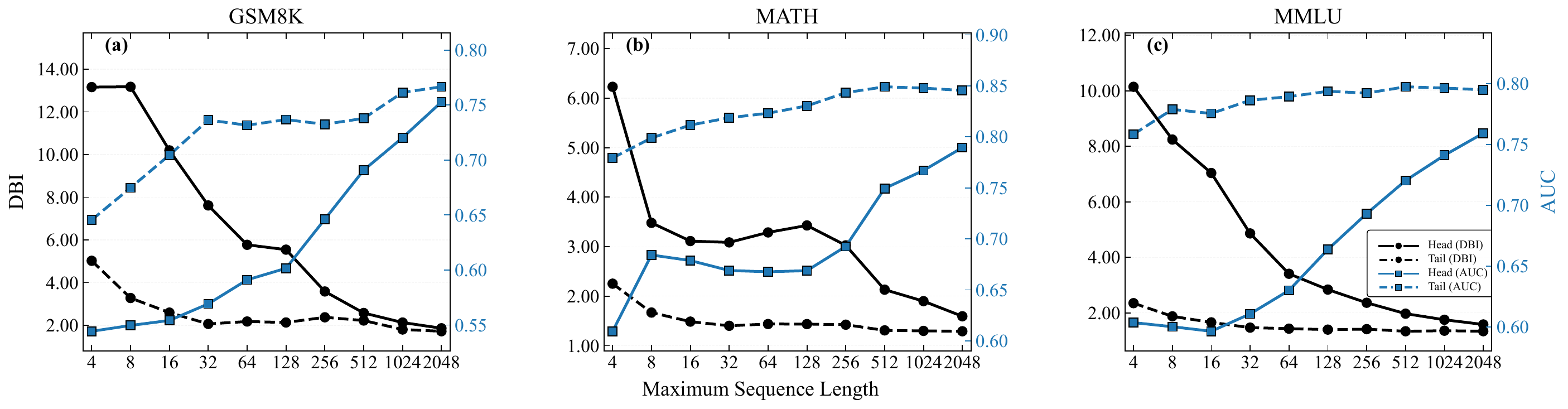}
    \caption{
    Matched comparison between head-aligned and tail-aligned inputs across maximum input lengths on GSM8K, MATH and MMLU. For each $L_{\max}$, the head-aligned input retains the first $L_{\max}$ confidence values, whereas the tail-aligned input retains the last $L_{\max}$ confidence values. Tail-aligned inputs generally yield stronger class separation and higher trace-level discrimination.
    }
    \label{fig:head_tail}
\end{figure*}
\section{Matched Head-Versus-Tail Positional Comparison}
\label{app:head_tail}

The sliding-window analysis in the main text shows that windows closer to the end of the reasoning trace are generally more informative. To test this conclusion under a simpler matched comparison, we contrast head-aligned and tail-aligned inputs at identical maximum lengths.

For each length setting, we construct two inputs from the same trace. The head-aligned input retains the first $L_{\max}$ confidence values, whereas the tail-aligned input retains the last $L_{\max}$ confidence values. All other settings are unchanged. Across GSM8K, MATH and MMLU, tail-aligned inputs generally produce lower DBI and higher AUC than head-aligned inputs over a broad range of input lengths (Fig.~\ref{fig:head_tail}).

This result provides a controlled confirmation of the positional interpretation advanced in the main text. The stronger discrimination observed for tail-aligned inputs cannot be attributed simply to differences in input size, because head and tail are compared at matched lengths. Instead, it indicates that correctness-related structure is enriched near the end of reasoning, where the model's internal confidence dynamics may more directly reflect convergence, commitment or failure of the reasoning path.

\section{MMLU Subject List}
\label{app:mmlu_subjects}

Our MMLU subset consists of the following college- and high-school-level STEM subjects:
\begin{itemize}
    \item \texttt{college\_chemistry}
    \item \texttt{college\_mathematics}
    \item \texttt{college\_computer\_science}
    \item \texttt{college\_physics}
    \item \texttt{high\_school\_chemistry}
    \item \texttt{high\_school\_computer\_science}
    \item \texttt{high\_school\_mathematics}
    \item \texttt{high\_school\_physics}
    \item \texttt{high\_school\_statistics}
\end{itemize}

We focus on these subjects because they more often require analytical, quantitative and multi-step reasoning, making them more suitable for studying sampled reasoning traces and their associated confidence trajectories.

\bibliographystyle{IEEEtran}
\bibliography{references}

\end{document}